\documentclass[runningheads]{llncs}
\usepackage{graphicx}
\usepackage{booktabs}
\usepackage{multirow}
\usepackage{color}
\usepackage{comment}
\usepackage{courier}
\usepackage[linesnumbered,ruled,vlined]{algorithm2e}
\usepackage{amsmath} 

\usepackage{amssymb}
\usepackage{makecell}

\newcommand{\mc}[3]{\multicolumn{#1}{#2}{#3}}

\newcommand{\eg}{\textit{e.g.,~}}
\begin{document}
\title{DAWA: Dynamic Ambiguity-Wise Adaptation for Real-Time Domain Adaptive Semantic Segmentation}
\titlerunning{DAWA}
%

\author{Taorong Liu\inst{1} \and
Zhen Zhang\inst{1}\and
Liang Liao\inst{2}\and
Jing Xiao\inst{3}\thanks{Corresponding author} \and
Chia-Wen Lin\inst{4}
}
\authorrunning{Liu et al.}
%
\institute{School of Computer Science, Wuhan University\\
\email{cameltr@whu.edu.cn, zhangzhent@whu.edu.cn}\\
 \and
Hangzhou Institute of Technology, Xidian University\\
\email{liaoliang01@xidian.edu.cn}
\and
School of Artificial Intelligence, Wuhan University\\
\email{jing@whu.edu.cn}
 \and
Department of Electrical Engineering and the Institute of Communications Engineering, National Tsing Hua University\\
\email{cwlin@ee.nthu.edu.tw}
}

\maketitle              
\begin{abstract}
Test-time domain adaption (TTDA) for semantic segmentation aims to adapt a segmentation model trained on a source domain to a target domain for inference on-the-fly, where both efficiency and effectiveness are critical. However, existing TTDA methods either rely on costly frame-wise optimization or assume unrealistic domain shifts (\textit{e.g.}, night→snow), resulting in poor adaptation efficiency and continuous semantic ambiguities.
To address these challenges, we propose a real-time framework for TTDA semantic segmentation, called \textbf{Dynamic Ambiguity-Wise Adaptation (DAWA)}, which adaptively detects domain shifts and dynamically adjusts the learning strategies to mitigate continuous ambiguities in the test time. Specifically, we introduce the Dynamic Ambiguous Patch Mask (DAP Mask) strategy, which dynamically identifies and masks highly disturbed regions to prevent error accumulation in ambiguous classes. Furthermore, we present the Dynamic Ambiguous Class Mix (DAC Mix) strategy that leverages vision-language models to group semantically similar classes and augment the target domain with a meta-ambiguous class buffer. Extensive experiments on widely used TTDA benchmarks demonstrate that DAWA consistently outperforms state-of-the-art methods, while maintaining real-time inference speeds of approximately 40 FPS.

\end{abstract}
\section{Introduction}
\label{sec:intro}

\begin{figure}[t]
    \centering
    \includegraphics[width=0.8\linewidth]{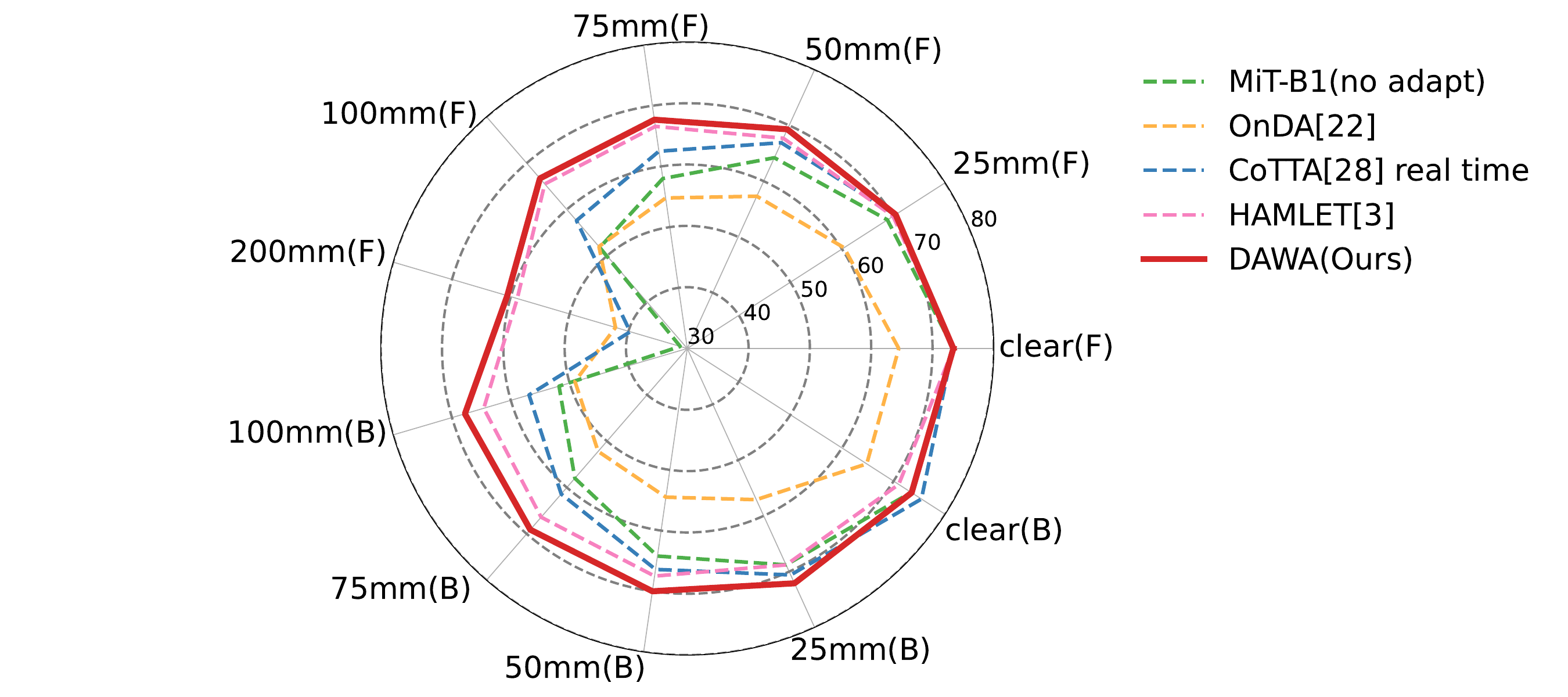} 
       \vspace{-2mm}
    \caption{Performance comparison of online adaptation methods on the Increasing Storm ~\cite{Panagiotakopoulos_ECCV_2022}. We plot mIoU(\%) achieved on each rainy intensity (\textit{i.e.}, clear, 25mm, 50mm, 75mm, 100mm, 200mm) of single domains. ``F'' represents forward adaptation (from \textit{clear} to 200mm), while ``B'' denotes backward adaptation (from 200mm to \textit{clear}).}
\label{fig:1}
   \vspace{-5mm}
\end{figure}

Semantic segmentation aims to annotate images at the pixel level to provide crucial information for various applications, especially in safety-critical fields such as autonomous driving. 
While state-of-the-art segmentation models~\cite{chen2017deeplab,xie2021segformer,jain2023oneformer} excel in clear weather due to abundant annotations, they often fail in dynamic environments that undergo rapid changes during practical driving, where the distribution of ongoing captured vision data continuously shifts due to factors like changing lighting and weather.
Unsupervised domain adaptation (UDA) has emerged as a promising approach for transferring learning-based models from labeled data distributions to unlabeled data. However, current UDA methods~\cite{bruggemann2023refign,hoyer2022hrda,li2023vblc,liaotip} are primarily tailored for static target domains and rely on offline training, making them unsuitable for handling continuous domain changes.

These challenges have led to the emergence of more demanding tasks, such as Test-Time Domain Adaptation (TTDA) \cite{wang2022continual,svdp,guo2025smoothing}, which involves rapidly fine-tuning models during test time to handle the changing data distribution in real-world scenarios. Despite its potential for on-site adaptation, TTDA faces significant limitations. For instance, continuously performing backpropagation on a per-frame basis incurs high computational costs and can reduce the overall framerate to meet the demands of continuous adaptation, resulting in a slower system. Additionally, optimizing on a per-frame basis for online adaptation could increase the risk of catastrophic forgetting of previous domains without enhancing adaptive performance. Furthermore, existing TTDA methods typically conduct adaptation on domain shift datasets \cite{sakaridis2021acdc} that deviate from realistic distributions (\eg day→night→rain→snow), which inadequately reflects real-world scenarios involving abrupt distribution shifts (\eg sudden rainstorms).

\begin{figure*}[t]
    \centering
    \includegraphics[width=\linewidth]{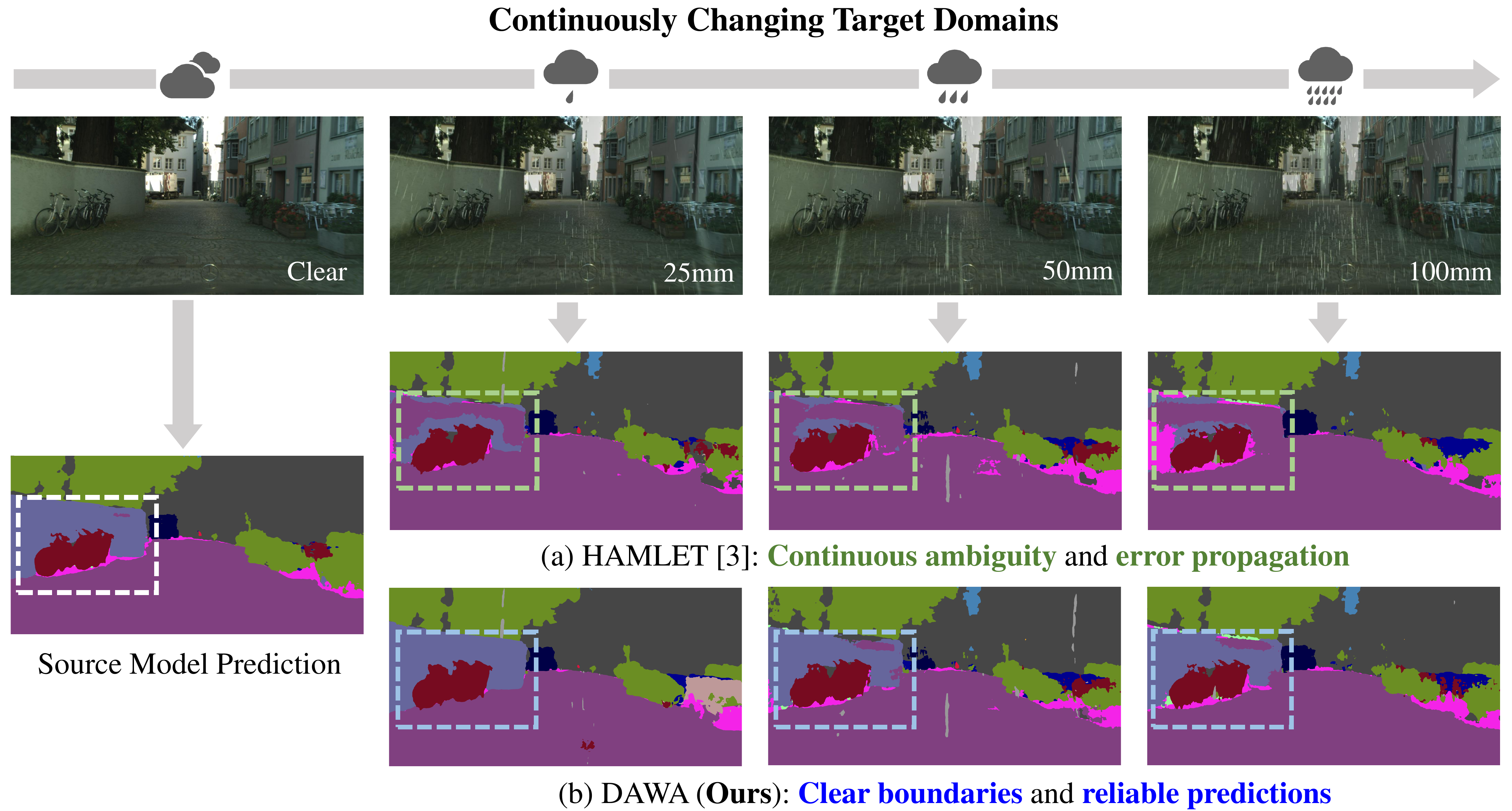}
       \vspace{-2mm}
    \caption{Challenge posed by visually similar classes under continuous adverse weather during online adaptation. 
    Under sudden rainfall, rain streaks cause certain classes (\textit{e.g.}, \textit{road} and \textit{wall}) to become easily ambiguous. While HAMLET~\cite{colomer2023hamlet} struggles with segmentation errors and error propagation under these conditions, our method robustly mitigates class ambiguity, yielding sharper boundaries and more reliable predictions.}
       \vspace{-4mm}
\label{motivation}
\end{figure*}

To address these issues, researchers have recently introduced Online Domain Adaptation (OnDA) \cite{Panagiotakopoulos_ECCV_2022,colomer2023hamlet}, which reduces computation by automatically detecting domain changes to trigger adaptation only when necessary. To mitigate catastrophic forgetting, OnDA employs a replay buffer to retain source knowledge while adapting to new domains. This paradigm leverages gradual environmental shifts during deployment without requiring pre-associated data. However, we observe that existing OnDA still struggle with continuous ambiguity among visually similar classes under dynamic adverse conditions. As depicted in Fig. \ref{motivation}, when encountering sudden rainfall, the state-of-the-art method, 
\textit{i.e.}, HAMLET~\cite{colomer2023hamlet}, makes consistent errors in distinguishing such classes (\eg \textit{road} and \textit{wall}). 
This suggests that the model transfers incorrect knowledge from prior domains, degrading performance in the current one. 

We identify two key factors underlying this issue:
(1) \textbf{Dynamic environmental disruptions} induce regional noise, blurring decision boundaries between ambiguous classes without explicit spatial suppression. This permits error propagation from noisy patches to dominate the learning;
(2) \textbf{Semantic ambiguity persistence} between visually similar classes is reinforced by continuous domain shifts, as existing methods rely on rigid class relationship assumptions that mismatch evolving semantic contexts. This leads to pseudo-label error accumulation due to random cross-domain augmentation.
Although approaches~\cite{Panagiotakopoulos_ECCV_2022,colomer2023hamlet} employ a domain indicator and adaptive self-training, their adaptation suffers from two critical limitations:
(1) Absence of spatial-aware noise suppression permits ambiguous patches to distort feature learning;
(2) Lack of semantic contextual relationship modeling fails to resolve persistent inter-class ambiguity, perpetuating a cycle of confirmation bias.

In this work, we propose Dynamic Ambiguity-Wise Adaptation (DAWA), a real-time framework for domain shift detection and model adaptation that explicitly addresses ambiguity and uncertainty arising from environmental disturbances and continuous distribution shifts.
The key idea is to enhance model robustness by focusing on ambiguous regions and classes that frequently occur in dynamic target domains. 
To this end, we introduce two core components: Dynamic Ambiguous Patch Mask (DAP Mask) and Dynamic Ambiguous Class Mix (DAC Mix).
Specifically, in DAP Mask, inspired by~\cite{zou2024freqmamba}, we analyze high-frequency energy distributions to identify semantically disturbed regions. A dynamic binary mask is generated to suppress these ambiguous patches, guiding the model to learn more stable representations from cleaner areas and mitigating error accumulation in uncertain classes. 
In DAC Mix, we leverage the semantic reasoning capabilities of Vision-Language Models (VLMs)~\cite{gpt4o,ma2024does} to construct a meta-ambiguous class buffer capturing semantically related class groups. Based on this buffer, we perform ambiguity-aware class mix, preserving contextual coherence and reducing confusion among visually similar classes. To summarize, our contributions are as follows:

(1) We propose Dynamic Ambiguity-Wise Adaptation (DAWA), a reliable framework tailored for TTDA semantic segmentation, which explicitly identifies and handles ambiguous regions and classes to enhance real-time adaptation.

(2) DAWA incorporates two novel modules to enhance adaptation: (i) DAP Mask leverages high-frequency energy analysis to identify highly disturbed regions and suppress error accumulation in ambiguous classes; and (ii) DAC Mix, a dynamic augmentation strategy guided by vision-language models that groups semantically related classes and augments the target domain with a meta-ambiguous class buffer to improve semantic disambiguation.

(3) Extensive experiments demonstrate the superior performance of our framework across various dynamic scenarios, achieving state-of-the-art results in segmentation accuracy while maintaining approximately 40 FPS.

\section{Related Work}
\noindent \textbf{UDA for Semantic Segmentation.}
 UDA aims to leverage labeled data from the source domain to develop a high-performing model for unlabeled data in the target domain. 
Common approaches include learning domain-invariant features via adversarial training~\cite{ganin}, style transfer~\cite{stylization,yang_fda_2020,icassp23}, similarity mining~\cite{tmmzhu}, and self-training with pseudo-labels~\cite{zou2019confidence,hoyer2022daformer}.
Recently, many researchers have explored the domain adaptation from \emph{normal} to \emph{adverse} weather~\cite{bruggemann2023refign,liao2023only,rizzoli2025cars}. It is highly relevant for practical scenarios such as automated driving.
However, these methods typically assume static target domains and rely on offline training, making them unsuitable for dynamic environments with continuous domain shifts.

\noindent \textbf{Test-Time Domain Adaptation.}
TTDA addresses adaptation during test-time deployment without access to source domain data. Common strategies include generating pseudo-source data~\cite{liu2021source}, freezing the final model layers~\cite{liang2020we}, entropy minimization~\cite{wang2021tent}, and prototypes adaptation~\cite{wang2023dynamically}.
A notable advancement is Continual Test-Time Adaptation (CoTTA)\cite{wang2022continual}, which adopts a weighted augmentation-averaged mean teacher framework and has inspired many follow-up works\cite{svdp,liu2024continual,lee2025dicotta}. However, these methods still perform adaptation on a per-frame basis, resulting in high adaptation cost. Furthermore, widely used TTDA benchmarks (e.g., ACDC~\cite{sakaridis2021acdc}, day→night→rain→snow) often diverge from real-world conditions, limiting the practicality of the results.

To address this issue, OnDA\cite{Panagiotakopoulos_ECCV_2022,colomer2023hamlet} has been proposed to focus on more realistic and complex scenarios. Unlike TTDA, OnDA handles continuous, unpredictable domain shifts without clear boundaries between domains.
OnDA\cite{Panagiotakopoulos_ECCV_2022} adopts self-training through coordinated static and dynamic teachers to enable online adaptation while mitigating forgetting, but at the cost of high computational overhead.
HAMLET~\cite{colomer2023hamlet} introduces a hardware-aware modular training framework to reduce time cost. However, it overlooks the challenge posed by visually similar classes affected by continuous domain scenes.

\begin{figure*}[t]
    \centering \includegraphics[width=\linewidth]{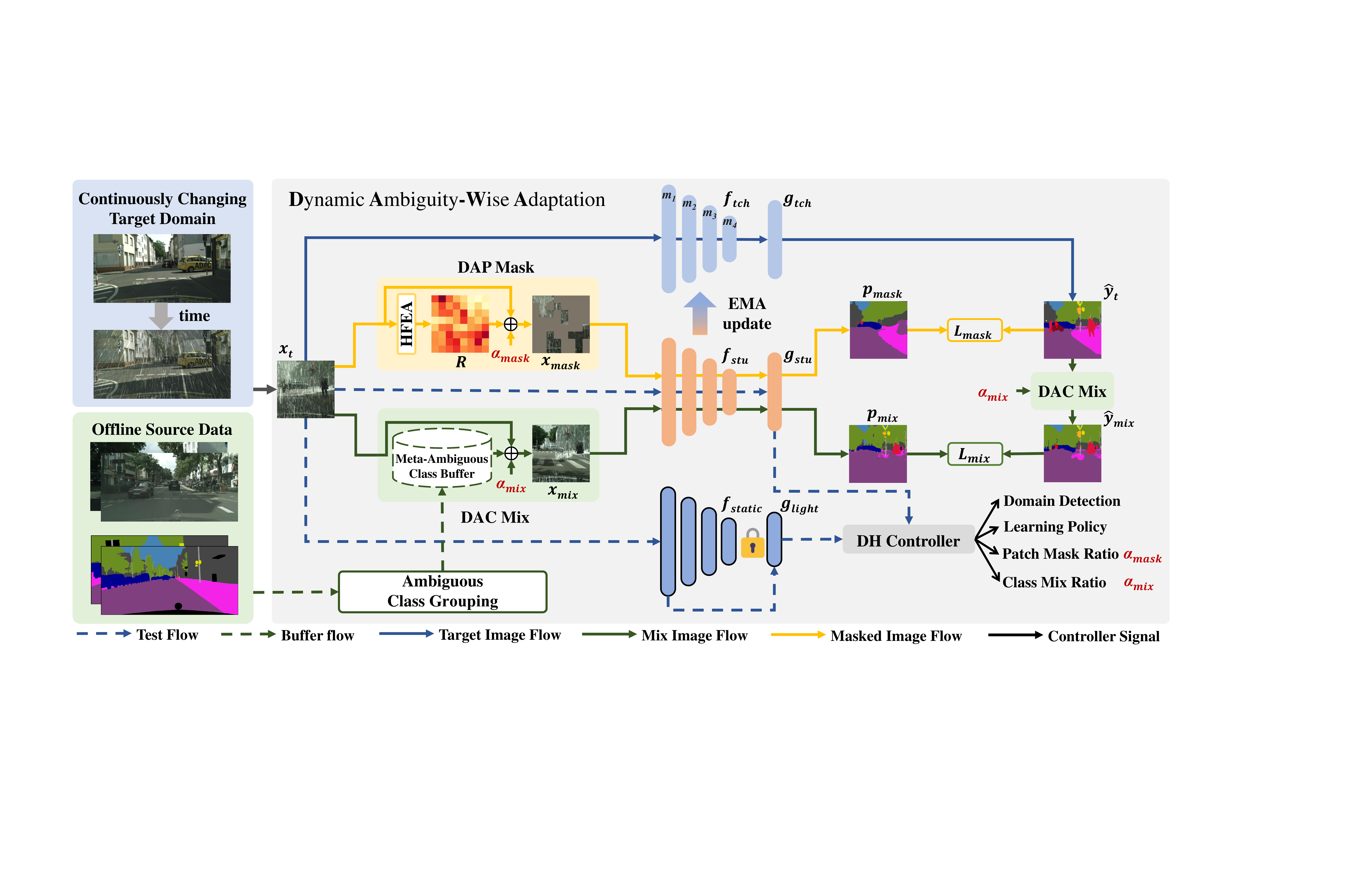}  
   \vspace{-2mm}
    \caption{The overview of our proposed DAWA. DAWA aims to detect and adapt to domain changes in real-time from the DH Controller, while simultaneously mitigating the ambiguity by DAP Mask and reducing the uncertainty by DAC Mix that arises from external environmental factors and continuous distribution shifts.}
   \vspace{-4mm}
\label{fig:framework}
\end{figure*}

\section{Proposed Method}

\subsection{Problem Formulation}
In TTDA for semantic segmentation, a pretrained segmentation model $f_\theta : \mathbf{x} \rightarrow y$ is deployed in an open-world setting with streaming test data from potentially shifting and unknown target domains. Unlike traditional TTDA, OnDA is to enable efficient adaptation by dynamically detecting domain transitions and updating parameters only when necessary. It addresses \textit{non-uniform}, \textit{discontinuous} temporal flows where each segment $t_i$ may belong to a distinct domain $\mathcal{D}_{t_i}$, reflecting realistic scenarios with implicit transitions and varying domain durations. Thus, OnDA must decide not only \textit{how} but also \textit{when} to adapt.

At each time segment $t_i$, the model $f_\theta$ processes a target sample $\mathbf{x}_{t_i}$ through three steps: (i) detecting domain changes via a signal-based mechanism, (ii) generating a prediction $\hat{y}_{t_i}$, and (iii) updating the model if stability is confirmed. In the next segment $t_{i+1}$, the model receives a new sample $\mathbf{x}_{t_{i+1}}$, potentially from a new domain $\mathcal{D}_{t_{i+1}}$ and without access to prior data, highlighting the online nature of the setting.

\subsection{Framework Overview}
We introduce \textbf{DAWA}, a reliable and real-time framework tailored for TTDA semantic segmentation.
As shown in Fig.~\ref{fig:framework}, DAWA detects and adapts domain shifts in real-time through the Dynamic Hyper-parameter (DH) controller, while simultaneously mitigating the ambiguity of visually similar classes by Dynamic Ambiguous Patch Mask (DAP Mask) and reducing the uncertainty and noisy label during training by Dynamic Ambiguous Class Mix (DAC Mix).

Specifically, DAWA consists of a teacher network $\phi_\text{tch}$, a student network $\phi_\text{stu}$, and a static teacher encoder $f_\text{static}$ paired with a lightweight decoder $g_\text{light}$. The network's backbone  $f$ is a modular encoder \cite{colomer2023hamlet} incorporating four distinct modules: $f = m_4 \circ m_3 \circ m_2 \circ m_1$.  
To handle unforeseen domain shifts, we employ a DH Controller that dynamically adjusts relevant hyper-parameters in response to detected changes and optimizes training efficiency. 
To improve the model's ability to handle ambiguous classes, we propose a DAP Mask strategy, which utilizes High-Frequency Energy Analysis (HFEA) to identify and emphasize ambiguous regions within scenes. 
Furthermore, DAC Mix is introduced to integrate a meta-ambiguous class buffer through the knowledge of the vision language model on the source data into the target scene to mitigate high uncertainty and noisy labels during online adaptation while reducing training costs and speeding up the adaptation process.
Finally, we leverage a masked loss $\mathcal{L}_\text{mask}$ to promote prediction consistency and a mixed loss $\mathcal{L}_\text{mix}$ to enhance training robustness, effectively handling continuous domain changes under adverse conditions.

\subsection{Dynamic Ambiguous Patch Mask}

Previous methods~\cite{hoyer2023mic} often use \textbf{\textit{random}} masking with a \textbf{\textit{fixed}} ratio, which  potentially neglects certain classes and introduce noise by ignoring scene-specific characteristics. However, the mask ratio should adapt to the evolving target domain. Inspired by~\cite{zou2024freqmamba}, we observe that high-frequency components indicate scene disturbance, where severe interference reduces model confidence in ambiguous classes. Therefore, we employ HFEA to measure disturbance levels and dynamically adjust the mask ratio based on domain-specific features, which reduces the impact of noisy regions and guides the model to focus on clearer areas, improving contextual understanding and prediction accuracy.

\noindent \textbf{High Frequency Energy Analysis.}
In Fourier spectral decomposition, low-frequency components represent smooth regions, while high-frequency components correspond to rapid intensity changes \cite{zou2024freqmamba} (\eg edges, textures, noise). High frequency is more focused on harsh areas, which are more prone to continuous class confusion segmentation errors across continuous domain shifts. Thus, we divide the target image into $N \times N$ grids $\mathbf{c}_{i,j}$ and estimate the disturbance level of the current scene using the Fast Fourier Transform to calculate the high-frequency energy ratio $R_{i,j}$ in a certain area. This process can be defined as: 
\begin{equation}
    R_{i,j} = \frac{\sum_{(u,v) \in \text{HF}} \log \left(|\text{FFT}(\mathbf{c}_{i,j})_{uv}| + \epsilon \right)}{\sum_{(u,v)} \log \left(|\text{FFT}(\mathbf{c}_{i,j})_{uv}| + \epsilon \right)},
\end{equation}

\begin{equation}
    \mathcal{M}_{i,j} = 
    \begin{cases}
    1 & \text{if } R_{i,j} \geq \text{percentile}(\mathcal{R}, \alpha_\text{mask}) \\
    0 & \text{otherwise}
    \end{cases},
    \end{equation}
where \(\text{FFT}(\mathbf{c}_{i,j})_{uv}\) denotes the value calculated by the Fast Fourier Transform of the grid $\mathbf{c}_{i,j}$ at the frequency coordinates \((u,v)\) in the frequency domain. $i$ and $j$ denote the rows and columns of the patch indices. \(\text{HF}\) represents the specified high-frequency region, with \((u,v)\) values indicating the range of high-frequency components. \(\epsilon\) is a small constant used to prevent negative infinity in logarithmic calculations. $\text{percentile}(\mathcal{R}, \alpha_\text{mask})$ refers to the value below which a given percentage $\alpha_\text{mask}$ of the $R_{i,j}$ in the set $\mathcal{R}$. After sorting the energy of certain regions, we generate masks for the top $\alpha_\text{mask}$ of all regions. $\alpha_\text{mask}$ is the ratio of the mask, whose value is determined by the DH Controller.

\noindent \textbf{Dynamic Mask Operation.}
 By element-wise multiplying the mask $\mathcal{M}$ from HFEA, we obtain the masked target image $x_\text{mask}$ with the target image $x_\text{t}$:
\begin{equation}
x_\text{mask} = \mathcal{M} \odot x_\text{t}.
\end{equation}
Therefore, the model is encouraged to learn contextual relationships of ambiguous classes from the continuously changing domains, providing additional cues for robust segmentation of classes with similar local appearances. 

\subsection{Dynamic Ambiguous Class Mix}

Effective data augmentation during online adaptation is critical for mitigating prediction uncertainty and model robustness. While ClassMix~\cite{olsson2021classmix} demonstrates promise through cross-domain class-wise mixing, it still suffer from \textit{static} mixing ratios and inadequate handling of semantic ambiguity in dynamic scenes. Meanwhile, unlike context-aware mix strategy~\cite{zhou2022context}, Our DAC Mix addresses this by leveraging Vision-Language Models (VLMs) to automatically discover\textbf{ semantic ambiguity} and eliminate manual heuristics while ensuring scalability.

\noindent \textbf{Ambiguous Class Grouping}
We leverage VLMs' exceptional capability in capturing high-level relationships between visual patterns and linguistic concepts to uncover structured ambiguity among classes. Specifically, GPT-4o~\cite{gpt4o} is employed not for manual heuristics but to automatically construct a hierarchical semantic structure of frequently confused classes in segmentation scenarios. Formally, these ambiguous class groups are defined as:
\begin{equation}
A_g = { C_1, C_2, ..., C_N },
\end{equation}
where\( A_g \)represents an ambiguous class group and \( C_n \)corresponds to its constituent fine-grained classes. This structured approach enables more effective ambiguity-aware data augmentation compared to conventional methods.

\noindent \textbf{Meta-Ambiguous Class Buffer}
Building on the semantic groupings, we design a Meta-Ambiguous Class Buffer for context-preserving augmentation. The process begins by randomly selecting an initial class from either domain's label map, then expanding it to include all contextually associated classes within its semantic group (e.g., \textit{Traffic Sign}, \textit{Traffic Light}, and \textit{Pole}). Following \cite{zhou2022context}, we generate a binary mask marking these classes' pixels as 1 (others as 0):

In this way, the Meta-Ambiguous Class Buffer stores a contextually extended class set reflecting semantic associations, along with a binary mask indicating their spatial distribution. This enables realistic mixing of class-wise content from the target domain, reinforcing the model’s understanding of ambiguous classes and promoting domain-consistent scene structures for better adaptation.

\noindent \textbf{Dynamic Mix Operation.}
In the mixing operation, the target pseudo labels \(p_\text{t}\) and source labels \(y_\text{s}\) are mixed based on a ratio \(\alpha_\text{mix}\). A binary mask \(\mathcal{M}_\text{mix}\) is used to determine which regions of the target image \(x_\text{t}\) are replaced by the source image \(x_\text{s}\), generating the mixed image \(x_\text{mix}\) and label \(\hat{y}_\text{mix}\) as follows:

\begin{equation}
x_\text{mix} = \mathcal{M}_\text{mix} \odot x_\text{s} + (1-\mathcal{M}_\text{mix}) \odot x_\text{t},
\end{equation}
\begin{equation}
\hat{y}_\text{mix} = \mathcal{M}_\text{mix} \odot y_\text{s} +  (1-\mathcal{M}_\text{mix}) \odot \hat{y}_\text{t},
\end{equation}
By including \(\alpha_\text{mix}\), the ratio between the source and target images can be flexibly adjusted to meet the needs of different scenarios.

\subsection{Dynamic Hyper-parameter Controller}
Unforeseeable changes in the target domain occur when the model encounters target data streams during deployment. Therefore, we extend \cite{colomer2023hamlet} to adjust the mask ratio \( \alpha_\text{mask} \) and mix ratio \( \alpha_\text{mix} \) dynamically during online adaptation.

\noindent \textbf{Domain Detection.}
To identify target domain shifts, we compute the cross-entropy loss between the predictions of $g_\text{stu}$ and $g_\text{light}$ as a measure of domain distance. This measurement of domain distance improves as the $g_\text{stu}$ continuously adapts over time. A domain shift is detected when the distance change exceeds a minimum threshold. We define $A_T^{(i)}$ as a denoised signal through bin-averaging calculated from the distance. Domains are modeled as discrete steps of $A_T^{(i)}$: 
\begin{equation}
B_0 = A_0, \qquad
B_i =
\begin{cases}
    A_i & \text{if  $|B_{i-1} - A_i|>z$} \\
    B_{i-1} & \text{otherwise}
\end{cases},
\end{equation}
where $B$ is the discretized signal and $z$ is the minimum distance used to identify new domains. A domain change is detected if the signed amplitude of the domain shifts $\Delta B_i = |B_i - B_{i-1}| > z$.

\noindent \textbf{Learning Policy.}
Upon detecting a domain shift, the number of adaptation iterations $L$ is determined by $L = K_l \frac{|\Delta B_i|}{z}$, where $|\Delta B_i|$ measures the shift magnitude relative to a threshold $z$, and $K_l$ is a scaling factor for adaptation intensity. The value of $K_l$ depends on the model’s proximity to the source domain:
\begin{equation}
K_l =
\begin{cases}
    K_l^\text{max} & \text{if } \Delta B_i \geq 0, \\
    K_l^\text{min} + \frac{(B_i - B_\text{source})(K_l^\text{max} - K_l^\text{min})}{B_\text{hard} - B_\text{source}} & \text{otherwise}
\end{cases},
\end{equation}
where $B_\text{source}$ and $B_\text{hard}$ denote the closest and furthest states from the source domain, respectively. $K_l^\text{min}$ and $K_l^\text{max}$ define the range of adaptation iterations.

The learning rate $\eta$ is linearly decayed during adaptation, with its initial value $K_\eta$ dynamically adjusted by:
\begin{equation}
K_\eta = K_\eta^\text{min} + \frac{(B_i - B_\text{source})(K_\eta^\text{max} - K_\eta^\text{min})}{B_\text{hard} - B_\text{source}}.
\end{equation}
Here, $K_\eta^\text{min}$ and $K_\eta^\text{max}$ correspond to easier and harder domain shifts, respectively, enabling adaptive learning based on shift severity.

\noindent \textbf{Patch Mask Ratio.}
We leverage domain knowledge to control the mask ratio based on the domain distance from the source domain during online adaptation. 
\begin{equation}
    \alpha_\text{mask} = \alpha_\text{mask}^\text{min} + \frac{(B_\text{i}-B_\text{source})(\alpha_\text{mask}^\text{max}-\alpha_\text{mask}^\text{min})}{B_\text{hard}-B_\text{source}},
\end{equation}
where $\alpha_\text{mask}^\text{min}$ is the value of $\alpha_\text{mask}$ assigned when the network is close to the source domain. 
and $\alpha_\text{mask}^\text{max}$ is, respectively, opposite in meaning to $\alpha_\text{mask}^\text{min}$.

\noindent \textbf{Class Mix Ratio.}
The class mix ratio during adaptation is also based on the domain knowledge:
\begin{equation}
    \alpha_\text{mix} = \alpha_\text{mix}^\text{min} + \frac{(B_\text{i}-B_\text{source})(\alpha_\text{mix}^\text{max}-\alpha_\text{mix}^\text{min})}{B_\text{hard}-B_\text{source}},
\end{equation}
where $\alpha_\text{mix}$ is the percentage of source classes used during adaptation, 
and $\alpha_\text{mix}^\text{min}$ is the value of $\alpha_\text{mix}$ assigned when the network is close to the source domain. 
and $\alpha_\text{mix}^\text{max}$ is, respectively, opposite in meaning to $\alpha_\text{mix}^\text{min}$.

\subsection{Overall Optimization}
Once the DH Controller detects domain changes and returns the training signal, the model enters the online training.

For $x_\text{mask}$, the masked prediction can only utilize limited information from the unmasked regions. We refer to \cite{hoyer2023mic} for consistency prediction constraints, utilizing mask consistency loss \(\mathcal{L}_\text{mask}\):
\begin{equation}
\mathcal{L}_\text{mask} = -\sum_{j=1}^{H \times W} \sum_{c=1}^{C} \hat{y}_{t} ^{(i,j,c)} \log \phi_\text{stu}(x_\text{mask}^{(i,j,c)}),
\end{equation}
where \(\hat{y}_\text{t}\) represent the quality weighted pseudo-labels predicted by the teacher network \(\phi_\text{tch}\) on the target image \(x_\text{t}\). Additionally, for the mixed image, we constrain the mixed pseudo-label loss $\mathcal{L}_\text{mix}$ using mixed pseudo-labels $\hat{y}_\text{mix}$ obtained from the DAC Mix: 
\begin{equation}
\mathcal{L}_\text{mix} = -\sum_{j=1}^{H \times W} \sum_{c=1}^{C} \hat{y}_\text{mix}^{(i,j,c)} \log \phi_\text{stu}(x_\text{mix}^{(i,j,c)}).
\end{equation}
The teacher network \(\phi_\text{tch}\) is implemented as an EMA teacher \cite{tarvainen2017mean}.  The total loss \(\mathcal{L}\) is : \(\mathcal{L} = \mathcal{L}_\text{mix} + \mathcal{L}_\text{mask}\).

\section{Experimental Results}

\begin{table*}[t]
    \centering
    \caption{Quantitative comparison against relevant methods on Increasing Storm~\cite{Panagiotakopoulos_ECCV_2022}. For each configuration, we report mIoU(\%) and framerate per second (FPS). Type ``T'' indicates the test time adaptation method, while ``O'' refers to the online adaptation method. The best two scores are indicated by \textbf{bold} and \underline{underline}.}
   \vspace{-2mm}
\resizebox{\linewidth}{!}
{
    \begin{tabular}{cl|c|cc cc cc cc cc c|ccc|cc}
    \toprule
    \mc{2}{c|}{\multirow{2}{*}{Methods}} &  \multirow{2}{*}{Type}  & \mc{2}{c}{clear} & \mc{2}{c}{25mm} & \mc{2}{c}{50mm} & \mc{2}{c}{75mm} & \mc{2}{c}{100mm} &  200mm &  \mc{3}{c|}{h-mIoU(\%)} &  \multirow{2}{*}{FPS}  \\
    & {} &      &  F &          B &      F &          B &          F &          B &      F &          B &          F &          B &      F &        F &  B &  All &  &  \\
    \midrule
    \midrule
    \mc{19}{c}{DeepLab-V2} \\
    \midrule
    (A) & No Adapt & --- & 64.5 & --- & 57.1 & --- & 48.7 & --- & 41.5 & --- & 34.4 & --- & 18.5 & 37.3 & --- & --- & 39.4 \\
    \midrule
    (B) & OnDA~\cite{Panagiotakopoulos_ECCV_2022} & O & 64.5 & 64.8 & 60.4 & 57.1 & 57.3 & 54.5 & 54.8 & 52.2 & 52.0 & 49.1 & 42.2 & 54.2 & 55.1 & 54.6 & 6.7 \\
    \midrule
    \midrule
    \mc{19}{c}{Segformer-B1} \\
    \midrule
    (C) & No Adapt & ---  &   73.4 & ---&  68.8 & ---&  64.2 & ---&  58.0 & ---&  51.8 & ---&  31.2 &      57.8 &  ---& ---&   60.8  \\ 
    \midrule
    (D) & TENT~\cite{wang2021tent} & T     &  73.0 &  72.8 &  68.5 &  68.6 &  64.5 &  64.8 &  59.7 &  60.2 &  54.5 &  54.8 &  35.9 &   56.2 &  63.6 &  59.9 &  10.0 \\
    (E) & TENT+Replay Buffer & T  &  73.0 &  72.8 &  68.5 &  68.6 &  64.5 &  64.8 &  59.7 &  60.2 &  54.4 &  54.7 &  35.8 &   56.1 &  63.6 &      59.9 &   7.8 \\
    (F) & CoTTA~\cite{wang2022continual} & T  &  72.5 & 74.4 & 69.5 & 70.9 & 65.9 &  68.2 & 66.1 & 64.7 & 64.6 & 63.5 & 57.2 & 65.6 &  68.1 & 66.8 & 0.6 \\
    (G) & CoTTA \textit{real-time} & T  & 73.3 & \bfseries 75.4 & 70.3 & 70.6 & 66.9 & 66.4 & 62.5 & 61.4 & 57.6 & 56.9 & 39.7 & 59.2 & 65.5 & 62.3 & 27.0  \\
    (H) & SVDP~\cite{svdp} & T   &  \bfseries 75.8 &  \underline{74.9} & \bfseries 71.4 & \underline{71.3}  & \underline{68.6}  &  \underline{69.3}  &   66.2	&  \underline{67.1} &  63.7 &   64.4 &  51.1  &  65.1  & \underline{69.2}  &  66.9  &   0.04  \\
    (I) & HAMLET~\cite{colomer2023hamlet} & O    &  \underline{73.4} &  71.0 & 69.9 &  68.8 &  67.7 &  67.5 &   \underline{66.6} &  66.4 &   \underline{65.5} &   \underline{64.6} &  \underline{58.9} &   \underline{66.7} &  67.6 &   \underline{67.1} &  \textbf{46.3}   \\
    \midrule
    (J) & DAWA (Ours) & O  &   \underline{73.4} &  73.5 &  \underline{70.4} &  \bfseries 71.6 & \bfseries 68.9 & \bfseries 70.0 &  \bfseries 67.6 & \bfseries 69.1 & \bfseries 66.1 &  \bfseries 66.4 &   \bfseries 60.6 & \bfseries  67.6 &  \bfseries 70.0 &  \bfseries 68.7 &  \underline{40.7}   \\
    \bottomrule
    \end{tabular}
}
   \vspace{-2mm}
    \label{tab:quantitative_comparison}
\end{table*}

\begin{table}[t]
\centering
\caption{Quantitative comparison on Increasing Fog~\cite{foggycityscapes}. We report mIoU(\%) and FPS. The best score is indicated by \textbf{bold}.}
   \vspace{-2mm}
\resizebox{0.8\linewidth}{!}{
\renewcommand\tabcolsep{4pt}
\begin{tabular}{l|ccccccc|ccc|ccc}
\toprule
\mc{1}{c|}{\multirow{2}{*}{Methods}}  & \mc{2}{c}{clear}  & \mc{2}{c}{600m} & \mc{2}{c}{300m} &  150m & \mc{3}{c|}{h-mIoU(\%)} &  \multirow{2}{*}{FPS} \\
{} &          F &          B &      F &          B &      F &          B &          F &         F & B & All &  \\
\midrule
(A) OnDA\cite{Panagiotakopoulos_ECCV_2022} & 64.9 & 65.4 & 63.8 & 62.7 & 61.7 & 59.5 & 51.6  & 60.0 & 62.4 & 61.0& 6.4 \\
\midrule
(B) HAMLET\cite{colomer2023hamlet} &  73.4	& 72.8 & 	71.3 & 	71.3 & 	69.1	& 69.5 & 	66	& 69.8 	& 71.2 	& 70.4 	& \textbf{42.0} \\
(C) DAWA &  \textbf{73.4} & \textbf{73.3}  & \textbf{71.6} &	\textbf{72.0} &	\textbf{69.9} &	\textbf{70.0} &	\textbf{66.6} &	\textbf{70.3} & 	\textbf{71.6}  &	\textbf{70.9}  & 38.6 \\
\bottomrule
\end{tabular}
}
   \vspace{-2mm}
\label{tab:fog}
\end{table}

\subsection{Experimental Settings}
\noindent \textbf{Datasets.}
We set Cityscapes dataset \cite{Cityscapes} under \textit{clear} weather conditions as the source domain. The experiments are carried out on the Increasing Storm~\cite{Panagiotakopoulos_ECCV_2022} and Increasing Foggy Cityscapes ~\cite{foggycityscapes}, which are a semi-synthetic benchmark with 2,975 training images and 500 validation images and apply synthetic rain and fog at different intensities and visibilities. The main benchmark, Increasing Storm~\cite{Panagiotakopoulos_ECCV_2022}, presents a pyramidal intensity profile. We selected rainy domains with the intensity of \{25mm, 50mm, 75mm, 100mm, 200mm\} and foggy domains with \{600m, 300m, and 150m\} visibility as experimental domains.

\noindent \textbf{Implementation Details.}
The online models were trained using AdamW with $\beta_1$ = 0.9, $\beta_2$ = 0.999, and weight decay 0.01. Following \cite{colomer2023hamlet}, the hyper-parameters in the DH Controller are: $\alpha$ = 0.1, 
$K_l$ = 750, $K^\text{min}_{\eta}$ = $1.5 \times 10^{-4}$, $K^\text{max}_{\eta}$ = $6 \times 10^{-5}$, $K^\text{min}_{l}$ = 187, $K_l^\text{max}$ = 562, 
$\alpha^\text{min}_\text{mix}$ = 0.5, $\alpha^\text{max}_\text{mix}$ = 0.75, $\alpha^\text{min}_\text{mask}$ = 0.3 and $\alpha^\text{max}_\text{mask}$ = 0.7. We use $B_\text{source}$ = 0.8, $B_\text{hard}$ = 2.55 for the rainy and foggy scenarios. 
During the DAP Mask stage, the mask patch size is 64, and \(\epsilon\) is set to $1 \times 10^{-7}$ to prevent negative infinity in logarithmic calculations. The evaluation metrics are mIoU(\%), FPS, and the harmonic mean (h-mIoU(\%)) over domains to present the overall adaptation performance. All models were trained on an RTX 4090 GPU.

\begin{figure*}[t]
	\centering
		\begin{tabular}{c}
	\includegraphics[width=0.95\linewidth]{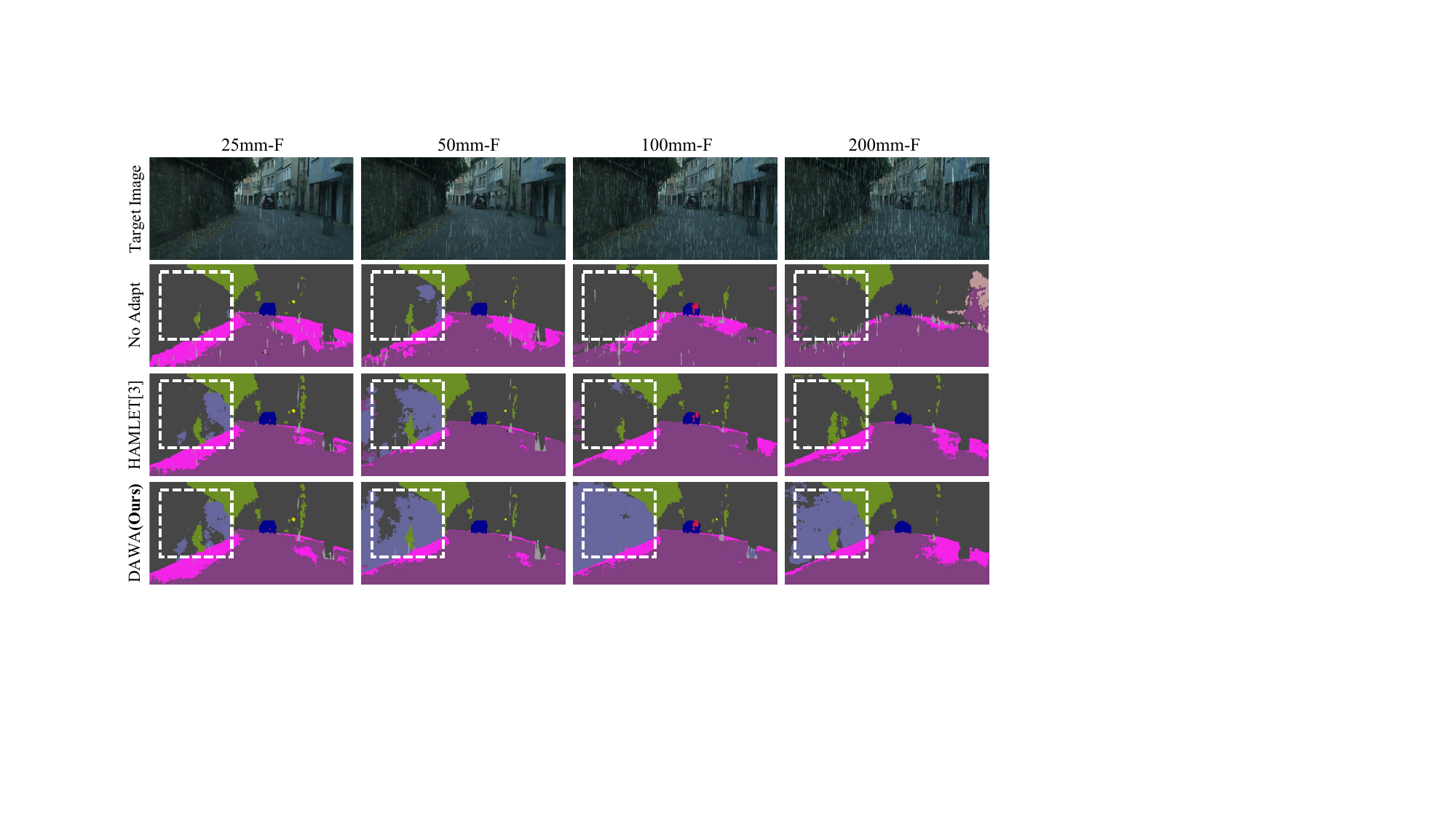} \\
    (a) Increasing Storm\cite {Panagiotakopoulos_ECCV_2022} \\
    \includegraphics[width=0.95\linewidth]{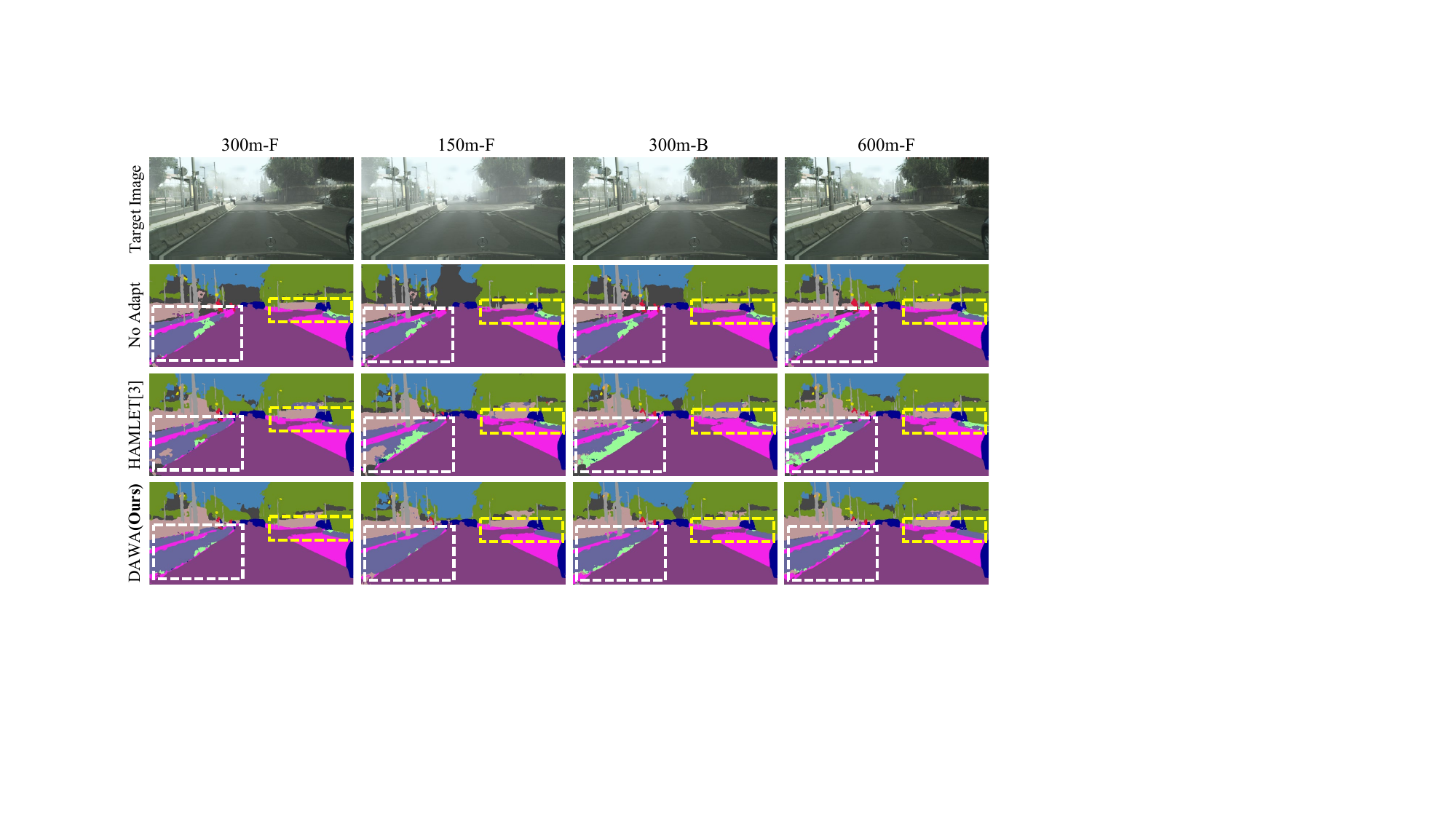} \\
    (b) Increasing Fog \cite{foggycityscapes} \\
	\end{tabular}
\vspace{-3mm}
   \caption{Qualitative comparisons on (a) Increasing Storm\cite {Panagiotakopoulos_ECCV_2022} and (b) Increasing Fog \cite{foggycityscapes} among source model (No Adapt), HAMLET\cite{colomer2023hamlet} and DAWA.}
   \vspace{-2mm}
\label{fig:qualitative}
\end{figure*}

\subsection{Performance Comparison}

\noindent \textbf{Quantitative Comparison.}
Tab.~\ref{tab:quantitative_comparison} and Tab.~\ref{tab:fog} provides a comparison of \textbf{DAWA} with other relevant methods under Increasing Storm and Increasing Fog. Columns ``F'' indicate forward adaptation from \textit{clear} to 200mm, while columns ``B'' show backward adaptation. Columns ``All'' show the all adaptation. The h-mIoU refers to the overall harmonic mean. Methods are categorized into different backbones. Type ``T'' and ``O'' indicates test time adaptation and online adaptation. 

Our proposed DAWA (J) emerges as a standout method, particularly in its handling of backward adaptation scenarios. Unlike other methods, DAWA not only avoids catastrophic forgetting but also improves performance as it focuses on ambiguous regions. This is evident in its consistent performance in backward scenarios, where it surpasses even its forward adaptation results. With an FPS of 40, DAWA achieves an excellent balance between accuracy and efficiency, making it a strong candidate for real-time applications.

\noindent \textbf{Qualitative Comparison.}
Fig.\ref{fig:qualitative} present continuous qualitative comparisons in dynamic target domains, featuring results from the pre-trained model, HAMLET, and DAWA.
In rainy scenes (Fig.\ref{fig:qualitative}(a)), the pre-trained model struggles due to its clear-weather training, and while HAMLET improves upon it, it still misclassifies ambiguous classes like \textit{wall} and \textit{building}. DAWA, by contrast, better exploits contextual cues to yield more accurate results.
Under foggy conditions (Fig.\ref{fig:qualitative}(b)), HAMLET has difficulty with classes like \textit{terrain} and \textit{wall}, whereas DAWA maintains stronger performance. However, both methods fail in extreme fog where distant objects are severely obscured.
Overall, DAWA shows superior adaptability to adverse weather by mitigating continuous ambiguities. 

\begin{table*}[t]
\centering
\caption{Ablation studies of the effect of components on Increasing Storm~\cite{Panagiotakopoulos_ECCV_2022}. For each configuration, we report the mIoU(\%) and FPS. Best scores are indicated by \textbf{bold}.}
\vspace{-3mm}
\resizebox{\linewidth}{!}{
\renewcommand\tabcolsep{2pt}
\begin{tabular}{ccc|cc cc cc cc cc c|ccc|cc}
\toprule
{} & \mc{2}{c|}{Components}  & \mc{2}{c}{clear} & \mc{2}{c}{25mm} & \mc{2}{c}{50mm} & \mc{2}{c}{75mm} & \mc{2}{c}{100mm} &  200mm &  \mc{3}{c|}{h-mIoU(\%)} &  \multirow{2}{*}{FPS}  \\
  
  & DAC Mask & DAC Mix &  F &          B &      F &          B &          F &          B &      F &          B &          F &          B &      F &        F &  B &  All &  &  \\
\midrule
(A) & \mc{2}{c|}{No Adapt}  & 73.4 & ---&  68.8 & ---&  64.2 & ---&  58.0 & ---&  51.8 & ---&  31.2 &  57.8 &  ---& ---&   60.8     \\
\midrule
(B) & \checkmark & --  &  73.4 & 	68.7 &	69.2 &	67.4 &	66.8 &	66.2 &	64.3 &	63.8 &	62.5&	63.6&	56.8&	65.1& 	65.9 &	65.4  & 37.5 \\
(C) & -- & \checkmark  & 73.4  & 73.0 & 70.1 &  71.0  & 67.5 & 68.5  & 66.6  & 67.3 & 64.6 & 65.1 & 59.4 & 66.6 & 68.9 & 67.6 & \bfseries 45.0   \\
(D) & \checkmark & \checkmark &  \bfseries 73.4 &  \bfseries 73.5 &  \bfseries 70.4 &  \bfseries 71.6 & \bfseries 68.9 & \bfseries 70.0 &  \bfseries 67.6 & \bfseries 69.1 & \bfseries 66.1 &  \bfseries 66.4&   \bfseries 60.6 & \bfseries  67.6 &  \bfseries 70.0 &  \bfseries 68.7 &  40.7    \\
\bottomrule
\end{tabular}
}
   \vspace{-2mm}
\label{tab:ablation}
\end{table*}

\begin{figure*}[t]
    \centering
    \includegraphics[width=\linewidth]{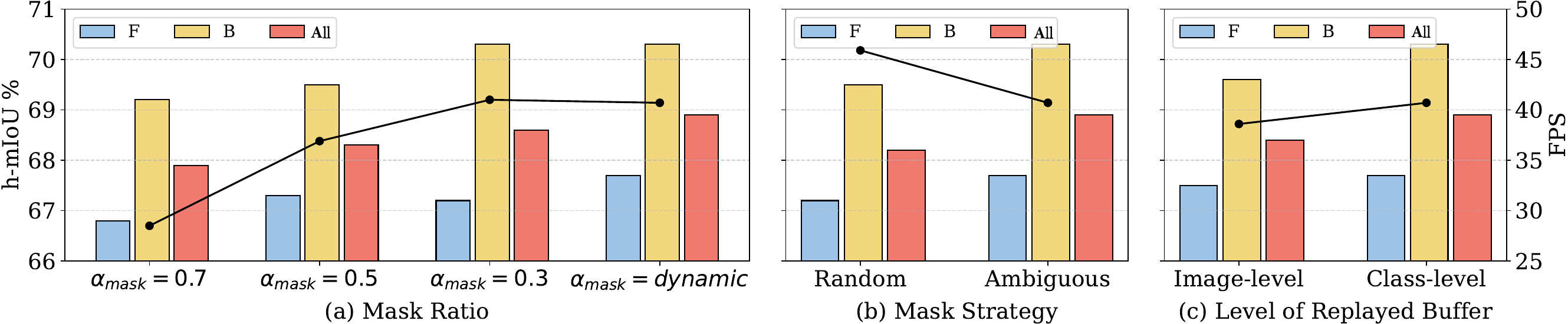}
    \vspace{-4mm}
    \caption{Ablation studies on (a) mask ratio, (b) mask strategy, and (c) replayed buffer. We report FPS (black line) and h-mIoU(\%) for forward (F), backward (B), and total (T) adaptation.}
       \vspace{-4mm}
\label{fig:ablation}
\end{figure*}

\subsection{Ablation Study}
\noindent \textbf{Effect of Components.}
Tab.~\ref{tab:ablation} presents each component's contribution under the Increasing Storm~\cite{Panagiotakopoulos_ECCV_2022}. While the source model (A) shows limited performance, independently adding DAP Mask (B) or DAC Mix (C) improves mIoU with moderate FPS reduction. (D) combines both components, achieving the best overall performance with over 40 FPS. Despite a small FPS drop, the trade-off remains acceptable, demonstrating that DAP Mask and DAC Mix effectively balance accuracy and efficiency for practical deployment.

\noindent \textbf{Design of DAP Mask.} 
To evaluate the DAP Mask, we study the effect of mask ratio and masking strategy. Fig.\ref{fig:ablation}(a) shows that fixed mask ratios yield suboptimal results, while the dynamic ratio delivers the best performance. In Fig.\ref{fig:ablation}(b), ambiguous masking on easily confused areas significantly outperforms random masking. In conclusion, These results confirm the effectiveness of the DAP Mask in enhancing adaptation by focusing on ambiguous regions.

\noindent \textbf{Levels of Replayed Buffer.}
We also conducted experiments to demonstrate the superiority of our proposed meta-ambiguous class buffer. As shown in Fig.\ref{fig:ablation} (c), the class-level (meta-ambiguous) buffer achieves higher FPS and better accuracy. This is because image-level replay can introduce excessive interference with current learning. Overall, the meta-ambiguous class buffer proves more suitable for online domain adaptation.

\begin{table}[t]
\centering
\caption{Comparison of different mix strategies on Increasing Storm. We report mIoU(\%) of 200mm, all mIou and FPS. Best scores are indicated by \textbf{bold}.}
   \vspace{-2mm}
\resizebox{0.8\linewidth}{!}{
\renewcommand\tabcolsep{4pt}
\begin{tabular}{lcccc}
\toprule
\textbf{Methods} & \textbf{Main Idea} & \textbf{200mm} & \textbf{All} & \textbf{FPS} \\
\midrule
ClassMix~\cite{olsson2021classmix}      & Random Augmentation   & 59.1 & 67.5 & \textbf{41.0} \\
CAMix~\cite{zhou2022context}         & Context Awareness     & 60.2 & 68.3 & 40.6 \\
DAC Mix (Ours)          & Semantic Ambiguity    & \textbf{60.6} & \textbf{68.7} & 40.7 \\
\bottomrule
\end{tabular}
}
\label{tab:mix_strategy}
\vspace{-2mm}
\end{table}

\noindent \textbf{Different Mix Strategies.}
As shown in Tab.\ref{tab:mix_strategy}, our DAC Mix outperforms random mix \cite{olsson2021classmix} and context-aware mix \cite{zhou2022context} methods, achieving the highest All mIoU of 68.7 while maintaining a competitive FPS of 40.7. This highlights the superiority of semantic ambiguity guidance in online adaptation.

\section{Conclusion}
In this study, we present the Dynamic Ambiguity-Wise Adaptation (DAWA) for solving real-time domain adaptive semantic segmentation under adverse weather conditions. DAWA address the challenge of ambiguous classes by continuous domain shifts that arise in online environments with DAP Mask and DAC Mix.
Experiments on OnDA benchmarks show that DAWA achieves state-of-the-art performance, highlighting its competitiveness. 
Future work may explore extending these strategies to other complex scenarios, further improving its adaptability and performance.
\newline

\noindent \textbf{Acknowledgement.} This work was supported by the National Natural Science Foundation of China (62371351, 62202349), the Young Elite Scientists Sponsorship Program by CAST (2023QNRC001),  and the Fundamental Research Funds for the Central Universities (ZYTS25036).

%
%
%
\bibliographystyle{splncs04}
\bibliography{reference}

@inproceedings{bruggemann2023refign,
  title={Refign: Align and refine for adaptation of semantic segmentation to adverse conditions},
  author={Br{\"u}ggemann, David and Sakaridis, Christos and Truong, Prune and Van Gool, Luc},
  booktitle={WACV},
  pages={3174--3184},
  year={2023}
}

@article{chen2017deeplab,
  title={Deeplab: Semantic image segmentation with deep convolutional nets, atrous convolution, and fully connected crfs},
  author={Chen, Liang-Chieh and Papandreou, George and Kokkinos, Iasonas and Murphy, Kevin and Yuille, Alan L},
  journal={IEEE Trans. Pattern Anal. Mach. Intell.},
  volume={40},
  number={4},
  pages={834--848},
  year={2017},
  publisher={IEEE}
}

@inproceedings{Cityscapes,
  title={The cityscapes dataset for semantic urban scene understanding},
  author={Cordts, Marius and Omran, Mohamed and Ramos, Sebastian and Rehfeld, Timo and Enzweiler, Markus and Benenson, Rodrigo and Franke, Uwe and Roth, Stefan and Schiele, Bernt},
  booktitle={CVPR},
  pages={3213--3223},
  year={2016}
}

@ARTICLE{tmmzhu,
  author={Zhu, Huilin and Yuan, Jingling and Zhong, Xian and Liao, Liang and Wang, Zheng},
  journal={IEEE Trans. Multimedia}, 
  title={Find Gold in Sand: Fine-Grained Similarity Mining for Domain-Adaptive Crowd Counting}, 
  year={2024},
  volume={26},
  number={},
  pages={3842-3855},}

@ARTICLE{liaotip,
  author={Liao, Liang and Chen, Wenyi and Xiao, Jing and Wang, Zheng and Lin, Chia-Wen and Satoh, Shin’ichi},
  journal={IEEE Trans. Image Process.}, 
  title={Unsupervised Foggy Scene Understanding via Self Spatial-Temporal Label Diffusion}, 
  year={2022},
  volume={31},
  number={},
  pages={3525-3540},}

@INPROCEEDINGS{icassp23,
  author={Zhong, Xian and Li, Wei and Liao, Liang and Xiao, Jing and Liu, Wenxuan and Huang, Wenxin and Wang, Zheng},
  booktitle={ICASSP}, 
  title={Bat: Bi-Alignment Based On Transformation in Multi-Target Domain Adaptation for Semantic Segmentation}, 
  year={2023},
  volume={},
  number={},
  pages={1-5},}

@article{stylization,
  title={Domain stylization: A fast covariance matching framework towards domain adaptation},
  author={Dundar, Aysegul and Liu, Ming-Yu and Yu, Zhiding and Wang, Ting-Chun and Zedlewski, John and Kautz, Jan},
  journal={IEEE Trans. Pattern Anal. Mach. Intell.},
  volume={43},
  number={7},
  pages={2360--2372},
  year={2020},
}

@article{ganin,
  title={Domain-adversarial training of neural networks},
  author={Ganin, Yaroslav and Ustinova, Evgeniya and Ajakan, Hana and Germain, Pascal and Larochelle, Hugo and Laviolette, Fran{\c{c}}ois and March, Mario and Lempitsky, Victor},
  journal={Journal of machine learning research},
  volume={17},
  number={59},
  pages={1--35},
  year={2016}
}

@inproceedings{guo2025smoothing,
  title={Smoothing the shift: Towards stable test-time adaptation under complex multimodal noises},
  author={Guo, Zirun and Jin, Tao},
  booktitle={ICLR‌},
  year={2025}
}

@inproceedings{hoyer2022daformer,
  title={Daformer: Improving network architectures and training strategies for domain-adaptive semantic segmentation},
  author={Hoyer, Lukas and Dai, Dengxin and Van Gool, Luc},
  booktitle={CVPR},
  pages={9924--9935},
  year={2022}
}

@inproceedings{hoyer2022hrda,
  title={HRDA: Context-Aware High-Resolution Domain-Adaptive Semantic Segmentation},
  author={Hoyer, Lukas and Dai, Dengxin and Van Gool, Luc},
  booktitle = {ECCV },
  year = {2022},
  pages={372--391},
}

@inproceedings{olsson2021classmix,
  title={Classmix: Segmentation-based data augmentation for semi-supervised learning},
  author={Olsson, Viktor and Tranheden, Wilhelm and Pinto, Juliano and Svensson, Lennart},
  booktitle={WACV},
  pages={1369--1378},
  year={2021}
}

@inproceedings{wang2022continual,
  title={Continual Test-Time Domain Adaptation},
  author={Wang, Qin and Fink, Olga and Van Gool, Luc and Dai, Dengxin},
  booktitle={Proceedings of Conference on Computer Vision and Pattern Recognition},
  year={2022}
}

@inproceedings{colomer2023hamlet,
  title={To adapt or not to adapt? real-time adaptation for semantic segmentation},
  author={Colomer, Marc Botet and Dovesi, Pier Luigi and Panagiotakopoulos, Theodoros and Carvalho, Joao Frederico and H{\"a}renstam-Nielsen, Linus and Azizpour, Hossein and Kjellstr{\"o}m, Hedvig and Cremers, Daniel and Poggi, Matteo},
  booktitle={ICCV},
  pages={16548--16559},
  year={2023}
}

@inproceedings{hoyer2023mic,
  title={MIC: Masked image consistency for context-enhanced domain adaptation},
  author={Hoyer, Lukas and Dai, Dengxin and Wang, Haoran and Van Gool, Luc},
  booktitle={CVPR},
  pages={11721--11732},
  year={2023}
}

@article{tarvainen2017mean,
  title={Mean teachers are better role models: Weight-averaged consistency targets improve semi-supervised deep learning results},
  author={Tarvainen, Antti and Valpola, Harri},
  journal={NIPS},
  volume={30},
  year={2017}
}

@inproceedings{sakaridis2021acdc,
  title={{ACDC}: The {Adverse} {Conditions} {Dataset} with {Correspondences} for semantic driving scene understanding},
  author={Sakaridis, Christos and Dai, Dengxin and Van Gool, Luc},
  booktitle={ICCV},
  year={2021}
}

@inproceedings{liao2023only,
  title={Only a few classes confusing: Pixel-wise candidate labels disambiguation for foggy scene understanding},
  author={Liao, Liang and Chen, Wenyi and Zhang, Zhen and Xiao, Jing and Yang, Yan and Lin, Chia-Wen and Satoh, Shin'ichi},
  booktitle={AAAI},
  volume={37},
  pages={1558--1567},
  year={2023}
}

@inproceedings{li2023vblc,
  title={VBLC: visibility boosting and logit-constraint learning for domain adaptive semantic segmentation under adverse conditions},
  author={Li, Mingjia and Xie, Binhui and Li, Shuang and Liu, Chi Harold and Cheng, Xinjing},
  booktitle={AAAI},
  volume={37},
  pages={8605--8613},
  year={2023}
}

@inproceedings{svdp,
  title={Exploring sparse visual prompt for domain adaptive dense prediction},
  author={Yang, Senqiao and Wu, Jiarui and Liu, Jiaming and Li, Xiaoqi and Zhang, Qizhe and Pan, Mingjie and Gan, Yulu and Chen, Zehui and Zhang, Shanghang},
  booktitle={AAAI},
  volume={38},
  pages={16334--16342},
  year={2024}
}

@article{foggycityscapes,
  author = {Sakaridis, Christos and Dai, Dengxin and Van Gool, Luc},
  title = {Semantic Foggy Scene Understanding with Synthetic Data},
  journal = {International Journal of Computer Vision},
  year = {2018},
  volume = {126},
  number = {9},
  pages = {973--992},
}

@inproceedings{zou2024freqmamba,
  title={FreqMamba: Viewing Mamba from a Frequency Perspective for Image Deraining},
  author={Zou, Zhen and Yu, Hu and Huang, Jie and Zhao, Feng},
  booktitle={ACM MM},
  year={2024}
}

@article{gpt4o,
  title={Gpt-4o system card},
  author={Hurst, Aaron and Lerer, Adam and Goucher, Adam P and Perelman, Adam and Ramesh, Aditya and Clark, Aidan and Ostrow, AJ and Welihinda, Akila and Hayes, Alan and Radford, Alec and others},
  journal={arXiv preprint arXiv:2410.21276},
  year={2024}
}

@article{zhou2022context,
  title={Context-aware mixup for domain adaptive semantic segmentation},
  author={Zhou, Qianyu and Feng, Zhengyang and Gu, Qiqi and Pang, Jiangmiao and Cheng, Guangliang and Lu, Xuequan and Shi, Jianping and Ma, Lizhuang},
  journal={IEEE Trans. Circuit Syst. Video Technol.},
  volume={33},
  number={2},
  pages={804--817},
  year={2022},
  publisher={IEEE}
}

@inproceedings{wang2023dynamically,
  title={Dynamically instance-guided adaptation: A backward-free approach for test-time domain adaptive semantic segmentation},
  author={Wang, Wei and Zhong, Zhun and Wang, Weijie and Chen, Xi and Ling, Charles and Wang, Boyu and Sebe, Nicu},
  booktitle={CVPR},
  pages={24090--24099},
  year={2023}
}

@inproceedings{liu2024continual,
  title={Continual-mae: Adaptive distribution masked autoencoders for continual test-time adaptation},
  author={Liu, Jiaming and Xu, Ran and Yang, Senqiao and Zhang, Renrui and Zhang, Qizhe and Chen, Zehui and Guo, Yandong and Zhang, Shanghang},
  booktitle={CVPR},
  pages={28653--28663},
  year={2024}
}

@inproceedings{jain2023oneformer,
  title={Oneformer: One transformer to rule universal image segmentation},
  author={Jain, Jitesh and Li, Jiachen and Chiu, Mang Tik and Hassani, Ali and Orlov, Nikita and Shi, Humphrey},
  booktitle={CVPR},
  pages={2989--2998},
  year={2023}
}

@article{lee2025dicotta,
  title={DiCoTTA: Domain-invariant Learning for Continual Test-time Adaptation},
  author={Lee, Sohyun and Kim, Nayeong and Kang, Juwon and Oh, Seong Joon and Kwak, Suha},
  journal={arXiv preprint arXiv:2504.04981},
  year={2025}
}

@inproceedings{ma2024does,
    title={Does VLM Classification Benefit from LLM Description Semantics?},
    author={Pingchuan Ma and Lennart Rietdorf and Dmytro Kotovenko and Vincent Tao Hu and Björn Ommer},
      booktitle={AAAI},
      year={2025},
}

@inproceedings{rizzoli2025cars,
  title={When cars meet drones: Hyperbolic federated learning for source-free domain adaptation in adverse weather},
  author={Rizzoli, Giulia and Caligiuri, Matteo and Shenaj, Donald and Barbato, Francesco and Zanuttigh, Pietro},
  booktitle={WACV},
  pages={1587--1596},
  year={2025},
}

@article{xie2021segformer,
  title={SegFormer: Simple and efficient design for semantic segmentation with transformers},
  author={Xie, Enze and Wang, Wenhai and Yu, Zhiding and Anandkumar, Anima and Alvarez, Jose M and Luo, Ping},
  journal={NIPS},
  volume={34},
  pages={12077--12090},
  year={2021}
}

@inproceedings{Panagiotakopoulos_ECCV_2022,
  title     = {Online Domain Adaptation for Semantic Segmentation in Ever-Changing Conditions},
  author    = {Panagiotakopoulos, Theodoros and
               Dovesi, Pier Luigi and
               H{\"a}renstam-Nielsen, Linus and
               Poggi, Matteo},
  booktitle = {ECCV},
pages={128--146},
  year = {2022}
}

@inproceedings{yang_fda_2020,
title = {{FDA}: Fourier Domain Adaptation for Semantic Segmentation},
pages = {4084--4094},
booktitle = {CVPR},
publisher = {{IEEE}},
author = {Yang, Yanchao and Soatto, Stefano},
year = {2020},
}

@inproceedings{zou2019confidence,
  title={Confidence regularized self-training},
  author={Zou, Yang and Yu, Zhiding and Liu, Xiaofeng and Kumar, BVK and Wang, Jinsong},
  booktitle={ICCV},
  pages={5982--5991},
  year={2019}
}

@inproceedings{liu2021source,
  title={Source-free domain adaptation for semantic segmentation},
  author={Liu, Yuang and Zhang, Wei and Wang, Jun},
  booktitle={CVPR},
  pages={1215--1224},
  year={2021}
}

@inproceedings{liang2020we,
  title={Do we really need to access the source data? source hypothesis transfer for unsupervised domain adaptation},
  author={Liang, Jian and Hu, Dapeng and Feng, Jiashi},
  booktitle={ICML},
  year={2020},
}

@inproceedings{wang2021tent,
  title={Tent: Fully Test-Time Adaptation by Entropy Minimization},
  author={Wang, Dequan and Shelhamer, Evan and Liu, Shaoteng and Olshausen, Bruno and Darrell, Trevor},
  booktitle={ICLR},
  year={2021}
}
%




\end{document}